\documentclass[10pt,twocolumn,letterpaper]{article}

\usepackage{cvpr}              
\definecolor{cvprblue}{rgb}{0.21,0.49,0.74}
\usepackage[pagebackref,breaklinks,colorlinks,allcolors=cvprblue]{hyperref}
\usepackage{soul}
\usepackage{float}
\usepackage{subcaption}
\usepackage{booktabs}
\usepackage{multirow}
\usepackage{siunitx}  

\def\paperID{*****} 
\def\confName{CVPR}
\def\confYear{2026}

\title{Fall Risk and Gait Analysis in Community-Dwelling Older Adults using World-Spaced 3D Human Mesh Recovery}

\author{
Chitra Banarjee\textsuperscript{1} \quad
Patrick Kwon\textsuperscript{2} \quad
Ania Lipat\textsuperscript{3} \quad
Rui Xie\textsuperscript{4} \quad
Chen Chen\textsuperscript{2} \quad
Ladda Thiamwong\textsuperscript{3} \\[4pt]
\textsuperscript{1}College of Medicine, UCF \quad
\textsuperscript{2}Institute of AI, UCF \\[4pt]
\textsuperscript{3}College of Nursing, UCF \quad
\textsuperscript{4}College of Sciences, UCF \\[4pt]
{\tt\small chitra.banarjee@ucf.edu \quad
yo564250@ucf.edu \quad
ania.lipat@ucf.edu} \\
{\tt\small rui.xie@ucf.edu \quad
chen.chen@ucf.edu \quad
ladda.thiamwong@ucf.edu}
}

\begin{document}
\maketitle
\begin{abstract}
Gait assessment is a key clinical indicator of fall risk and overall health in older adults. However, standard clinical practice is largely limited to stopwatch-measured gait speed. We present a pipeline that leverages a 3D Human Mesh Recovery (HMR) model to extract gait parameters from recordings of older adults completing the Timed Up and Go (TUG) test. From videos recorded across different community centers, we extract and analyze spatiotemporal gait parameters, including step time, sit-to-stand duration, and step length. We found that video-derived step time was significantly correlated with IMU-based insole measurements. Using linear mixed effects models, we confirmed that shorter, more variable step lengths and longer sit-to-stand durations were predicted by higher self-rated fall risk and fear of falling. These findings demonstrate that our pipeline can enable accessible and ecologically valid gait analysis in community settings.
\end{abstract}    
\section{Introduction}
\label{sec:intro}

Movement is central to human interaction and well-being across the lifespan \cite{cuignet2020, buchman2009, fogel1992}. Especially for older adults, movement is an indicator of overall health in the physical, mental, and cognitive domains \cite{delacamara2020, chang2022, studenski2019}. As such, gait evaluation is a key clinical measure for identifying older adults at risk for several physiological and pathological aging conditions, commonly used in primary care visits and geriatric assessments to screen participants at risk for frailty or falls \cite{lee2017, seematter2018}.

However, clinical gait evaluation remains largely limited to stopwatch-measured gait speed, as comprehensive assessments are constrained by limited access to technology and specialized training. Although the biomechanical correlates of fall risk are well-established, existing methodologies, such as wearable inertial sensors, optoelectronic marker-based systems, and multi-camera markerless motion capture, require dedicated infrastructure and considerable technical training, restricting their deployment beyond controlled clinical or research settings.

To solve this, we present a novel computer vision-informed approach that evaluates 3D gait measures using a Human Mesh Recovery (HMR) method, from a single monocular video camera that records the Timed Up and Go (TUG) test for older adults. While prior studies relied on 2D keypoints \cite{cao2019, fang2022, toshev2014} or multi-camera setups, we leverage Ground View HMR (GVHMR) \cite{Shen2024WorldGroundedHM} as the foundation of our pipeline to recover full human body motion anchored in \textit{world coordinates}. Unlike camera-relative methods that conflate human and camera motion, GVHMR reconstructs the participant's true trajectory through space, enabling the extraction of spatiotemporal gait parameters in \textit{absolute metric units}. Also, since older adult populations are sensitive to environments where unfamiliar equipment can limit participation, our approach is well-suited for deployment in real-world community settings.

Using GVHMR, we extract spatiotemporal variables from videos of older adults completing the TUG taken across several community and senior centers, which are compared to in-shoe pressure insoles to evaluate validity. We also explore how measures of fall risk predict various aspects of gait, and evaluate the use of HMR methods to analyze movement in community settings. 

Our main contributions are summarized as follows:
\begin{itemize}
    \item A gait acquisition protocol that uses a single camera to record older adults completing the TUG test across multiple community centers, only requiring limited clinical infrastructure and supervision.

    \item A novel application of world-grounded 3D HMR to clinical gait analysis, leveraging GVHMR to reconstruct full-body trajectories and spatiotemporal gait parameters.

    \item An automated pipeline extracting spatiotemporal parameters of gait and TUG transitions via signal processing and peak detection.

    \item A correlation analysis comparing video-derived step time against insole measurements, and a linear mixed effects (LME) framework evaluating how fall risk factors predict spatiotemporal gait in community-dwelling older adults.
\end{itemize}

\section{Related Work}
\label{sec:relatedworks}

Several studies have demonstrated the relevance of gait metrics for fall risk in older adults. For instance, \cite{bytyci2021, kim2025} showed that shorter step length and increased step width variability are associated with adverse outcomes, while pressure-based measures have been linked to both fall risk and cognitive impairment \cite{banarjee2025, akter2025, mawarikado2025}. Beyond spatial measures, segmentation of subtasks of the TUG test (e.g. standing up and turning) has demonstrated that increased turn duration and sit-to-stand times are associated with higher fall risk \cite{luis2026, lofgren2026, ortega2023}.

Recently, computer vision has been used for various healthcare applications, for both medical images \cite{javaid2024, khang2024, elyan2022, friedrich2025, yan2025} and human recordings \cite{jakubowski2017, li2022, ranjan2025, salcedo2024, shen2024}. With the advent of deep learning, human pose estimation methods have greatly evolved over the past decades \cite{marr2010, zheng2023}, advancing research in medical fields such as gait analysis and sports optimization \cite{ren2025, lu2024, mroz2021}. Several works in gait analysis further explored how these methods can increase the accessibility of fall risk assessment in labs and assisted living centers \cite{blasco2024, ng2020, luo2018, hou2021, won2020}, though deployment in uncontrolled community environments remains limited \cite{zhu2025, delfi2021}. These few studies in community settings rely on 2D skeleton-based models \cite{cao2019, fang2022, toshev2014, xu2022vitpose}, which cannot recover depth information or separate camera perspectives from human pose, making them ill-suited for extracting spatial measures such as step length. \ul{We aim to resolve this by using GVHMR \mbox{\cite{Shen2024WorldGroundedHM}}, which, unlike other pose estimation or human mesh recovery methods \mbox{\cite{Goel2023HumansI4, Kocabas2019VIBEVI, ye2023slahmr, Shin2023WHAMRW}}, efficiently predicts human poses in a \textit{gravity-aware world coordinate system}, enabling extraction of spatiotemporal gait parameters from a single monocular camera.}
\section{Methodology}
\label{sec:methods}

\begin{figure}
\vspace{-0.5cm}
    \centering
    \includegraphics[width=0.8\linewidth]{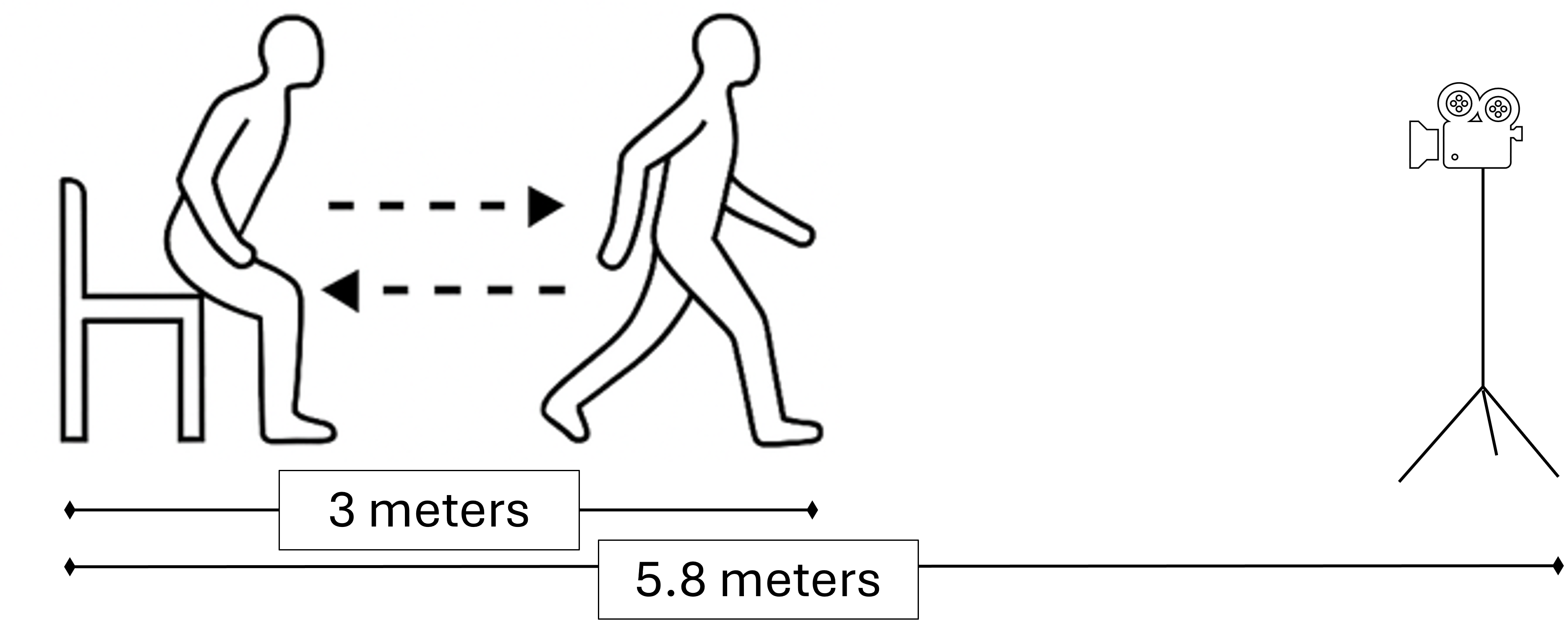}
    \vspace{-0.3cm}
    \caption{Schematic of Timed Up and Go (TUG) test set up}
    \label{fig:tug_setup}
    \vspace{-0.3cm}
\end{figure}

\begin{table*}
    \footnotesize
    \centering
    \begin{tabular}{p{3cm} p{4cm} p{9cm}}
        \hline
        \textbf{Variable} & \textbf{Assessment} & \textbf{Significance/Interpretation} \\
        \hline
        Self-Rated Fall Risk &
          Centers for Disease Control STEADI Score &
          12-item questionnaire quantifying self-rated fall risk, summed to produce the STEADI score, with higher scores indicating higher risk. Validated as indicator of fall risk \cite{bergen2019, nithman2019}. \\[2pt]
        Fear of Falling (FoF) &
          Short Falls Efficacy Scale-International (FES-I) Score &
          7-item questionnaire evaluating FoF, with higher scores indicating higher FoF. Validated as indicator of fall risk \cite{delbaere2010, kempen2008, yardley2005}. \\[2pt]
        Postural Sway &
          BTrackS Balance Test Score &
          Static balance assessment in which participants stand on a balance plate for 20 seconds three times to evaluate postural sway, with higher measures indicating worse static balance. Validated as indicator of fall risk \cite{levy2018, oconnor2016, goble2018}. \\
        \hline
    \end{tabular}
    \vspace{-0.3cm}
    \caption{The measures in this table were completed by the participants to provide information about subjective and objective fall risk, specifically self-rated fall risk, fear of falling, and postural sway.  }
    \label{tab:fall_risk}
    \vspace{-4mm}
\end{table*}

\subsection{Data Collection}
\textit{Video Timed Up and Go (vTUG).} Participants were instructed to complete the Timed Up and Go (TUG) test three times. The trials were recorded using a conventional monocular video camera, GoPro Hero 12, set up approximately 280~cm from the 3-meter mark (Fig.~\ref{fig:tug_setup}). The trials were then edited by trained URAs using Adobe Premiere Pro to trim and crop each video (30 fps).

\textit{Instrumented Timed Up and Go (iTUG).} In addition to the video, we used the wearable XSENSOR insole system \cite{parker2023} (XSENSOR Technology Corporation, Calgary, Canada, shown in \textbf{Appendix} \cref{fig:xsensor_pics}). The sensors are placed on the lateral malleoli of the ankle and collected data at 60 fps for accelerometer and gyroscope in three dimensions ($x$, $y$, $z$). These sensors served as a comparative reference for evaluating the video‑derived gait measures. While not a perfect ground truth, due to differences in coordinate systems, sampling rates, and our own processing choices, the insole data provides an independent sensor‑based benchmark to assess the temporal performance of the video pipeline.

\textit{Fall risk measures.} Participants were given questionnaires to evaluate subjective fall risk factors and a postural sway assessment to quantify balance (Table~\ref{tab:fall_risk}).

\begin{figure}[!t]
\vspace{-0.3cm}
    \centering
    \includegraphics[width=\linewidth]{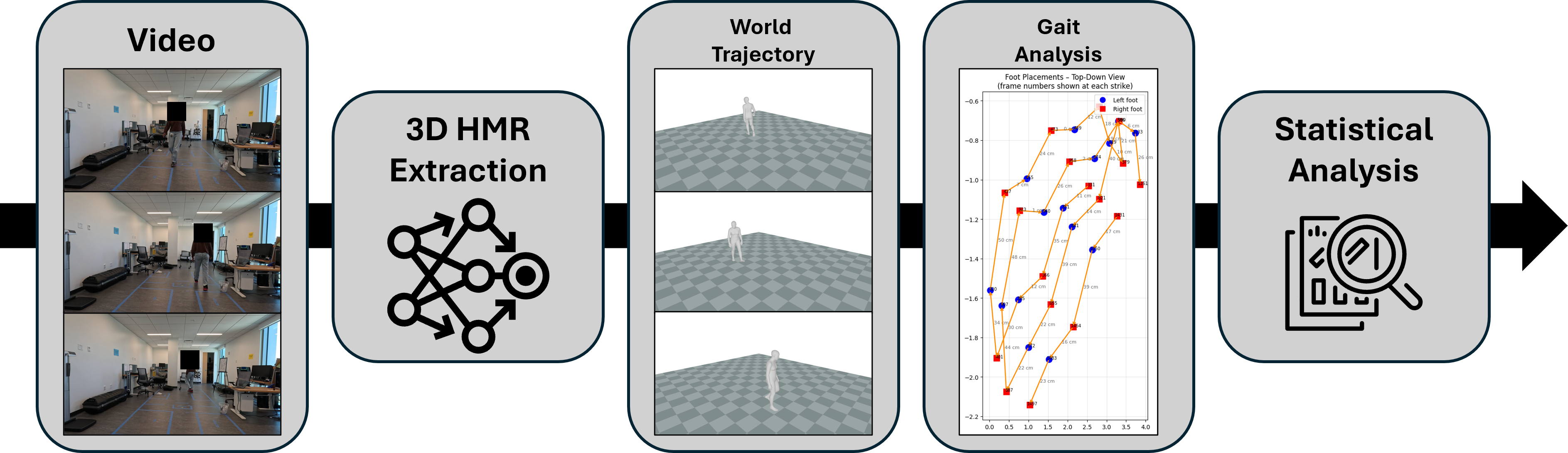}
    \vspace{-0.6cm}
    \caption{The overall pipeline of our study. Basing on the world trajectory extraction results from the TUG test using GVHMR \cite{Shen2024WorldGroundedHM} (Sec. 3.2), we analyze the gait factors (Sec. 3.3) and conduct statistical analysis (Sec. 3.4).}
    \label{fig:gvhmr}
    \vspace{-0.4cm}
\end{figure}
\subsection{Human Pose Estimation} The TUG (\cref{fig:tug_setup}) consists of standing up from a chair, walking, turning, and coming back to sit in the chair. Each TUG video is processed through GVHMR \cite{Shen2024WorldGroundedHM}, which recovers human pose separately for camera-space and world-space. Specifically, given a video $\{I^t\}_{t=0}^{T}$, GVHMR predicts the local body poses $\{\boldsymbol{\theta}^t \in \mathbb{R}^{21 \times 3}\}_{t=0}^{T}$, body shape coefficients $\boldsymbol{\beta} \in \mathbb{R}^{10}$ of SMPL-X \cite{Pavlakos2019ExpressiveBC}, the orientation $\{\boldsymbol{\Gamma}^t_c \in \mathbb{R}^{3}\}_{t=0}^{T}$ and translation $\{\boldsymbol{\tau}^t_cw \in \mathbb{R}^{3}\}_{t=0}^{T}$ corresponding to the camera-space trajectory, and the orientation $\{\boldsymbol{\Gamma}^t_w \in \mathbb{R}^{3}\}_{t=0}^{T}$ and translation $\{\boldsymbol{\tau}^t_w \in \mathbb{R}^{3}\}_{t=0}^{T}$ corresponding to the gravity-aligned world-space trajectory. Here, we mainly focus on the world-space trajectory information. The gait parameters are then derived from the 3D joint positions $\{\boldsymbol{J}^t \in \mathbb{R}^{24 \times 3}\}_{t=0}^{T}$, regressed from the SMPL-X kinematic model in world space. We present the TUG test setup in Fig.~\ref{fig:tug_setup}, and an example prediction result in Fig.~\ref{fig:gvhmr}. Further details are provided in \textbf{Appendix}.


\subsection{Gait Analysis}

Using the world-space joint trajectories recovered by GVHMR, we derive a set of spatiotemporal gait parameters that characterize participants’ mobility during the TUG test. These include temporal and spatial gait measures, and transition durations between subtasks (e.g., sit-to-stand and turning). The metrics are computed from the video-derived 3D joint trajectories, while wearable insole sensors are used as an independent reference to compare temporal gait measurements (Fig.~\ref{fig:gvhmr}).

Prior to analysis, we reduce high-frequency noise by applying Gaussian smoothing to each time series of 3D body keypoints from video data and to the wearable sensor signals. A discrete Gaussian kernel was constructed using the standard formula with $\sigma = 3$ samples. The kernel support was limited to $\pm 3\sigma$, resulting in a 19-point symmetric filter. The kernel was then normalized to unit sum prior to convolution.

\textit{Step Time (ST).} For the video-derived data, ST was calculated using GVHMR joint positions. For the insole-derived data, the raw data was exported through XSENSOR Pro Foot \& Gait System, and was calculated using peak detection with angular velocity (gyroscope $z$-axis) and vertical acceleration of ankles.

\textit{Step Length (SL) and Step Width (SW).} These were extracted by detecting the local minima of the GVHMR ankle joint positions to identify steps, then calculating the forward and lateral directions for SL and SW, respectively.

\textit{Sitting, Standing, Turning Transitions.} Similarly, peak detection for sit-to-stand (STS-1) and stand-to-sit (STS-2) was computed by calculating a composite sit-to-stand (STS, Eq.~\ref{eq:sts}) and hip line signal (Eq.~\ref{eq:hipline}), which utilize the hip and shoulder coordinates to segment subtasks (Table~\ref{tab:signals}).

\vspace{-4mm}

\begin{equation}
\label{eq:sts}
\text{STS} = 1.0 \cdot \dot{y}_{\mathrm{hip}} + 0.7 \cdot \dot{z}_{\mathrm{shoulder}} + 0.5 \cdot \dot{\theta}_{\mathrm{trunk}}
\end{equation}

\vspace{-6mm}

\begin{equation}
\label{eq:hipline}
\text{Hip Line Signal} = x_{R\,\mathrm{hip}} - x_{L\,\mathrm{hip}}
\end{equation}

\vspace{-2mm}

Further details are provided in \textbf{Appendix}.

\begin{table}
\footnotesize
\centering
\begin{tabular}{p{2.3cm} p{5.2cm}}
\hline
\textbf{Component} & \textbf{Indicator} \\
\hline
$\dot{y}_{\mathrm{hip}}$ &
  Vertical hip trajectory indicates initiation of STS, weighted the highest. \\
$\dot{z}_{\mathrm{shoulder}}$ &
  Anterior translation of the upper trunk during forward flexion; reliably marks the preparatory lean phase that precedes vertical rise. \\
$\dot{\theta}_{\mathrm{trunk}}$ &
  Angular trunk flexion--extension is noisier and more sensitive to small movements, thus assigned a lower weight. \\
Hip Line Signal &
  Alternating signed extrema in hip-line velocity reflect rapid angular deceleration and reversal of pelvic rotation during turning. \\
\hline
\end{tabular}
\vspace{-0.3cm}
\caption{Components of signals for segmentation of TUG subtask (sit-to-stand, turns, stand-to-sit).}
\label{tab:signals}
\vspace{-6mm}
\end{table}


\begin{figure}
    \centering
    \includegraphics[width=0.96\linewidth]{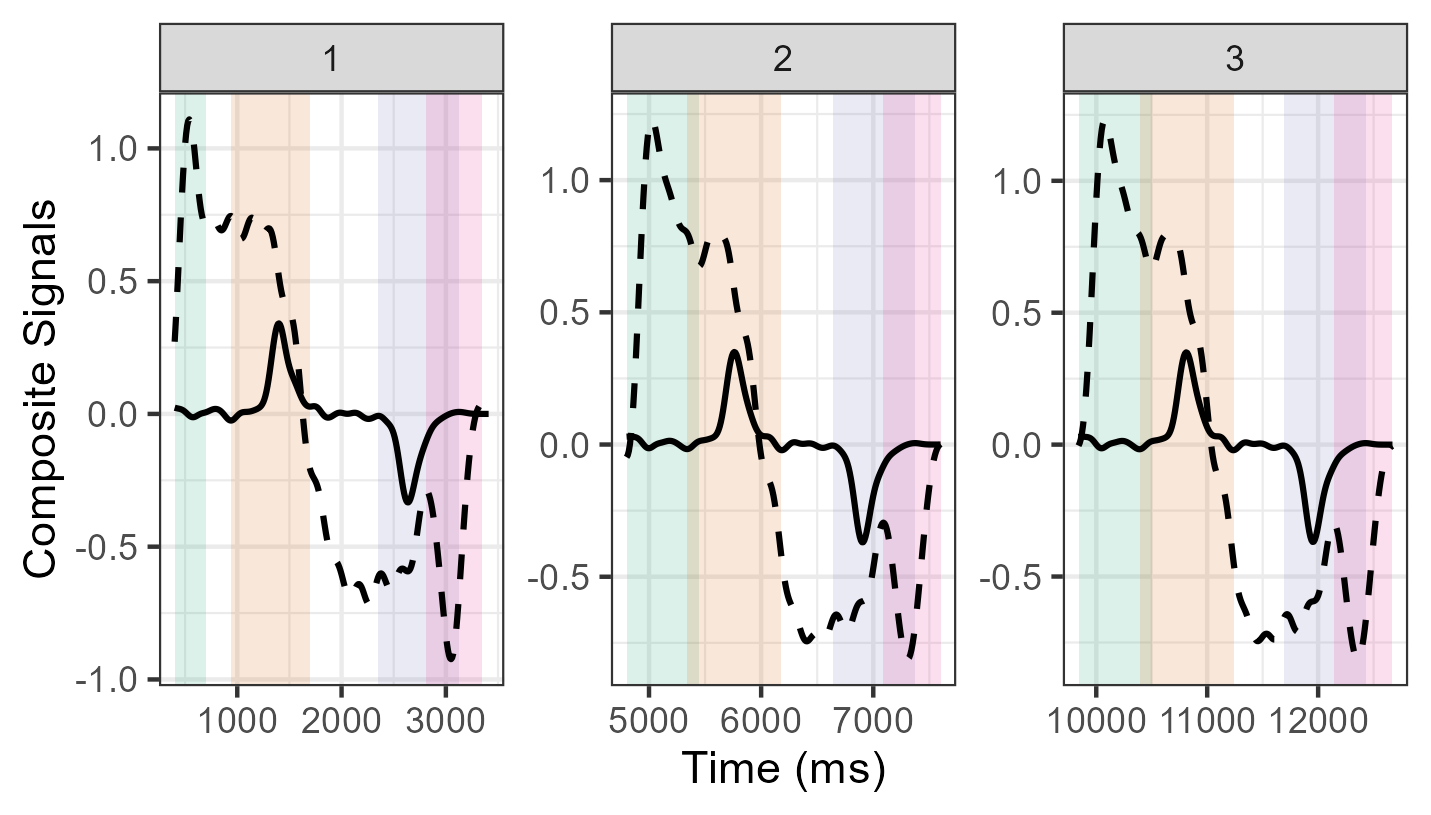}
    \vspace{-0.4cm}
    \caption{Automated Subtask Segmentation for three trials. Solid line refers to the hip line velocity and dashed line refers to sit-to-stand and stand-to-sit (STS) signal.}
    \label{fig:subtask}
    \vspace{-0.5cm}
\end{figure}

\subsection{Statistical Analysis}

To evaluate the reliability and clinical relevance of the extracted gait metrics, we conducted two sets of statistical analyses. First, we assessed the agreement between video-derived gait measures and wearable sensor measurements to validate the temporal accuracy of the video pipeline. Second, we examined whether the extracted gait parameters were associated with established fall risk indicators.

\textit{Correlation Analysis.} For this analysis, we aim to measure the correlation between ground-truth insole-derived sensor data and HMR-based video-derived data, to validate the usability of HMR. To improve the accuracy of this comparison and account for differences in each trial, only participants with three valid trials of the iTUG and vTUG were included for this analysis. The correlation analysis was conducted for each participant and trial from the 90 trials from 30 participants through a Spearman's correlation test after evaluation of normality using the Shapiro--Wilks test.

\textit{Linear Mixed Effects Models (LME).} Because each participant completed three trials of the TUG, the observations are not independent. As such, LMEs were used to model the data. Fixed effects consisted of fall risk factors (\cref{tab:fall_risk}) and age, with a random effect of participant to account for within-person variance (see  \cref{eq:lme}). Models predicted gait parameters (mean STS and durations, SL and SW mean, variability, and symmetry) for each trial. 

\section{Experiments}
\label{sec:experiments}

\subsection{Dataset Characteristics}
The final sample included 207 video recordings of the vTUG (video) from 52 older adults ($M_\text{age} = 74.2 + 7.81$ years), after removing participants without 3 valid trials. Participant demographics, including average self-rated fall risk, fear of falling, and postural control are shown in \cref{tab:demographics} of Appendix. Of those 207 trials, 30 assessments with 90 trials contained simultaneously collected iTUG. Further details are provided in \textbf{Appendix}.

\subsection{Insole and Video Comparison}

The Spearman's rank correlation analysis indicates moderate agreement between insole-derived and video-derived ST ($n=90, \rho=0.673, p < 0.001$). However, video-derived ST systematically underestimated the insole measurements.

\subsection{LME Analysis with Fall Risk Factors}

Self-rated fall risk (STEADI Score) significantly predicted STS duration. Turn durations were not predicted by any of the fall risk factors. Spatial measures of gait involved SL and SW, but only SL and SL variability were predicted by a fixed effect of self-rated fall risk (STEADI score) and fear of falling (FoF, short FES-I, Table 3). SW variability was predicted by age ($p < 0.01$) but not fall risk factors (\cref{tab:sw_variability}). The other measures, SW and symmetry measures, were not predicted by age or fall risk factors (Fig.~\ref{fig:subtask}, Table~\ref{tab:lme}). SL (\cref{tab:sl_full}) and SL Variability   (\cref{tab:sl_variability}) showed far stronger between‑participant consistency (ICC = 0.81) and a substantially better model fit (conditional R² = 0.85) than STS durations (\cref{tab:sts_durations}). Additional details about the statistical measures are described in \textbf{Appendix}.


\section{Discussion}
\label{sec:discussions}


We demonstrate the novel application of GVHMR to extract clinically meaningful gait metrics from monocular videos of older adults completing the TUG, with derived measures showing moderately strong correlations with simultaneously collected wearable-sensor data. Using LME models, we found that lower STEADI scores predicted longer sit-to-stand durations, though not turning durations (\cref{tab:sts_durations}). This result is in line with previous findings indicating a relationship between STS and fall risk. However, the null turning result contrasts with other literature that demonstrates a strong relationship with fall risk \cite{langhammer2006,luis2026, ortega2023}. Turn strategy (distinct or overlapping) varies widely between individuals, which may result in imprecise turn segmentation {\cite{weiss2016transition}. Future work can utilize GVHMR to evaluate turning strategy as an individual factor. Shorter and more variable step length was associated with greater self-rated fall risk and fear of falling (\cref{tab:sl_full}), consistent with literature linking cautious gait patterns to both perceived and physiological instability \cite{kim2025, brachman2025, marano2025}. Greater between-participant consistency and model fit indicate that gait characteristics such as SL may be more stable and more strongly tied to fall‑risk status than STS performance, which exhibited high within‑participant variability (\cref{tab:sl_full,tab:sts_durations}). These results suggest that older adults with poorer balance or reduced confidence tend to take smaller, more inconsistent steps and require more time to complete sit-to-stand transitions. These patterns may reflect compensatory strategies related to muscle weakness or fear of falling.


\begin{table}[t]
    \centering
    \footnotesize
    \begin{tabular}{l ccc}
    \hline
     & \multicolumn{3}{c}{\underline{STEADI Score}} \\
     & Estimate (CI) & & $p$ \\[1pt]
    \hline
    STS Duration & $1.23$ $(0.45, 2.01)$ & & $\mathbf{0.002}$ \\[4pt]
    Step Length (SL) & $-1.36$ $(-2.03, -0.68)$ & & $\mathbf{< 0.001}$ \\[4pt]
    SL Variability  & $-19.62$ $(-30.44, -8.80)$ & & $\mathbf{< 0.001}$ \\[4pt]
    \hline
     & \multicolumn{3}{c}{\underline{Short FES-I Score}} \\
    & Estimate (CI) & & $p$ \\
    \hline 
    STS Duration & $0.71$ $(-0.01, 1.42)$ & & $0.052$ \\[4pt]
    Step Length (SL) & $-1.04$ $(-1.65, -0.43)$  & & $\mathbf{0.001}$ \\[4pt]
    SL Variability & $-12.76$ $(-22.43, -3.09)$   & & $\mathbf{0.010}$ \\[4pt]
    \hline
    \end{tabular}
    \vspace{-0.2cm}
    \caption{Linear mixed effect model results of STEADI Score (self-rated fall risk) and short FES-I (fear of falling). ST refers to the video-derived step time. Estimate refers to the $\beta$ coefficient, CI = confidence interval, $p$ = $p$-value.}
    \label{tab:lme}
    \vspace{-0.2cm}
\end{table}

\begin{figure}
\vspace{-0.3cm}
    \centering
    \includegraphics[width=0.96\linewidth]{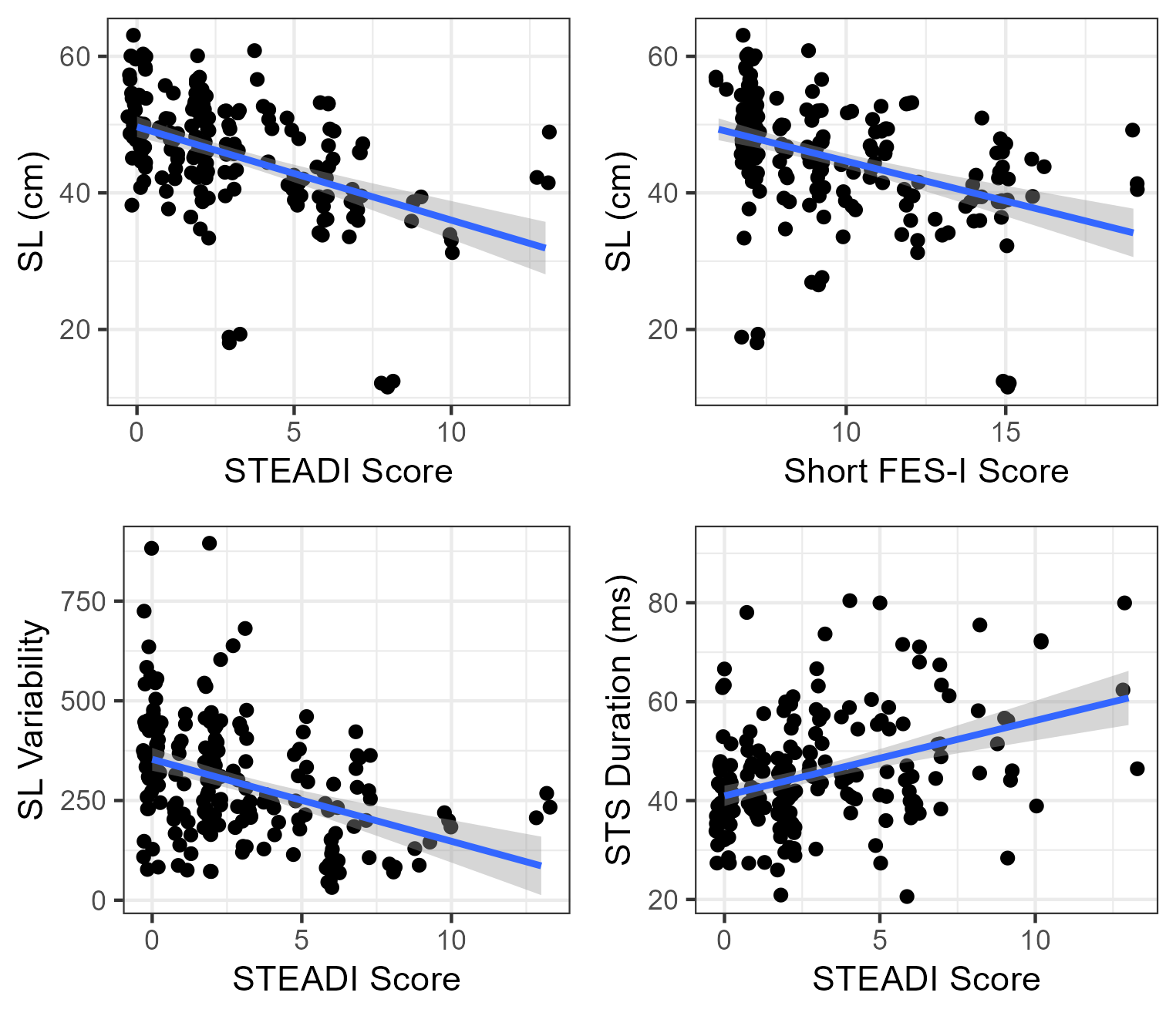}
    \vspace{-0.3cm}
    \caption{Scatterplots of mean step length (SL), SL variability, and sit-to-stand and stand-to-sit (STS) duration with fall risk factors (STEADI: self-rated fall risk, Short FES-I: fear of falling).}
    \label{fig:scatterplots}
    \vspace{-0.7cm}
\end{figure}

\vspace{-0.2cm}
\section{Conclusion}
\label{sec:conclusions} 

We offer a feasible and clinically relevant method for extracting gait and transitional movement metrics from a single TUG recording. GVHMR's ability to segment short, meaningful subtasks and produce measures that align with wearable-derived metrics highlights its potential as an accessible assessment tool. Associations between fall risk factors and specific gait features-particularly step length variability and sit-to-stand duration-underscore the value of video-based analysis for identifying subtle mobility changes. Future work should evaluate whether the derived metrics can enhance fall risk screening, support remote assessments, or predict prospective functional decline.


{
    \small
    \bibliographystyle{ieeenat_fullname}
    \bibliography{main}
}


\clearpage
\setcounter{page}{1}
\maketitlesupplementary

\section{Population Characteristics}
The sample consisted of community-dwelling older adults. Participants were recruited within Orlando, Florida using various strategies, including flyers, word-of-mouth, and collaboration with community partners. The inclusion criteria were that participants must be aged \(\geq\)60 years, be able to walk (with or without assistive devices but not requiring assistance from another person), live in their own homes or apartments, and be fluent in English or Spanish. The exclusion criteria were (1) having a medical condition that may preclude engagement in PA (including shortness of breath, dizziness, tightness or chest pain, and unusual fatigue at rest or with light exertion) and (2) currently receiving treatment from a rehabilitation facility. Participants provided written consent upon enrollment. All study procedures were approved by the University of Central Florida Institutional Review Board and all participants provided written informed consent prior to participation.

\begin{table}[h]
\centering
\footnotesize
\begin{tabular}{l c}
\toprule
\textbf{Variable} & \textbf{Mean $\pm$ Standard Deviation} \\
\midrule
Age                   & $74.2 \pm 7.81$ years   \\
STEADI Score          & $2.95 \pm 2.83$         \\
Short FES-I Score     & $9.40 \pm 3.02$         \\
BTrackS Balance Score & $29.3 \pm 11.2$ cm      \\
\bottomrule
\end{tabular}
\caption{Population characteristics.}
\label{tab:demographics}
\end{table}

\section{Specifics on GVHMR}

\textit{Comparisons to previous methods} Skeleton-based pose estimation methods such as OpenPose \cite{cao2019} and AlphaPose \cite{fang2022} extract two dimensional pixel-level joint positions from RGB images, which are fundamentally ill-suited for gait analysis; depth is unrecoverable from a single 2D projection, and measures such as step length and step width are conflated with camera perspective, rendering them neither metrically accurate nor view-invariant. While 3D HMR methods can reconstruct full-body geometry in three dimensions, most 3D HMR methods, such as HMR2.0 \cite{Goel2023HumansI4} and VIBE \cite{Kocabas2019VIBEVI} still estimate body pose only in the camera frame. Optimization-based methods, such as SLAHMR \cite{ye2023slahmr}, recover global trajectories but are computationally prohibitive and prone to convergence failures on long inputs, while regression-based alternatives, such as WHAM \cite{Shin2023WHAMRW}, require good initialization including a standardized environment with calibration to maintain consistent, accurate predictions. GVHMR \cite{Shen2024WorldGroundedHM} addresses these limitations by predicting pose in a per-frame \textit{Gravity-View coordinate system}, directly outputting gravity-aligned world-space trajectories without additional refinement networks, making it the most suitable method for processing extended, in-the-wild clinical video such as the TUG.

\textit{Calculation specifics} Each TUG video is processed through GVHMR \cite{Shen2024WorldGroundedHM}, which preprocesses each frame using human bounding boxes \cite{yolov8, Li2022CLIFFCL}, keypoints \cite{xu2022vitpose}, image features \cite{Goel2023HumansI4}, and estimated camera rotation \cite{Teed2022DeepPV}. These per-frame features are fused into tokens and passed through the model, recovering human pose within a \textbf{Gravity-View (GV) coordinate system} defined by the world gravity and camera view directions, which is naturally gravity-aligned and uniquely defined for each individual frame, reducing the ambiguity of the image-to-pose mapping. The model is based on a transformer architecture with Rotary Positional Embeddings (RoPE) \cite{Su2021RoFormerET}, which enables processing of sequences of arbitrary length without the need for a sliding window, making it suitable for application to TUG videos of varying lengths.

\begin{table}[t]
\centering
\begin{tabular}{lccc}
    \toprule
    Predictor & Estimate & CI & p \\
    \midrule
    (Intercept) & 296.47 & [160.02,\ 432.92] & $\mathbf{< 0.001}$ \\
    ste8        &  -0.14 & [-5.34,\ \ 5.05] & 0.956 \\
    age         &  -2.79 & [-4.73,\ \ 0.86] & $\mathbf{0.005}$ \\
    \bottomrule
\end{tabular}
\vspace{-2mm}
\caption{Step Width Variability}
\vspace{-2mm}
\label{tab:sw_model1}
\end{table}

\begin{figure}
\centering
    \begin{subfigure}[b]{0.4\linewidth}            
            \includegraphics[width=0.9\textwidth]{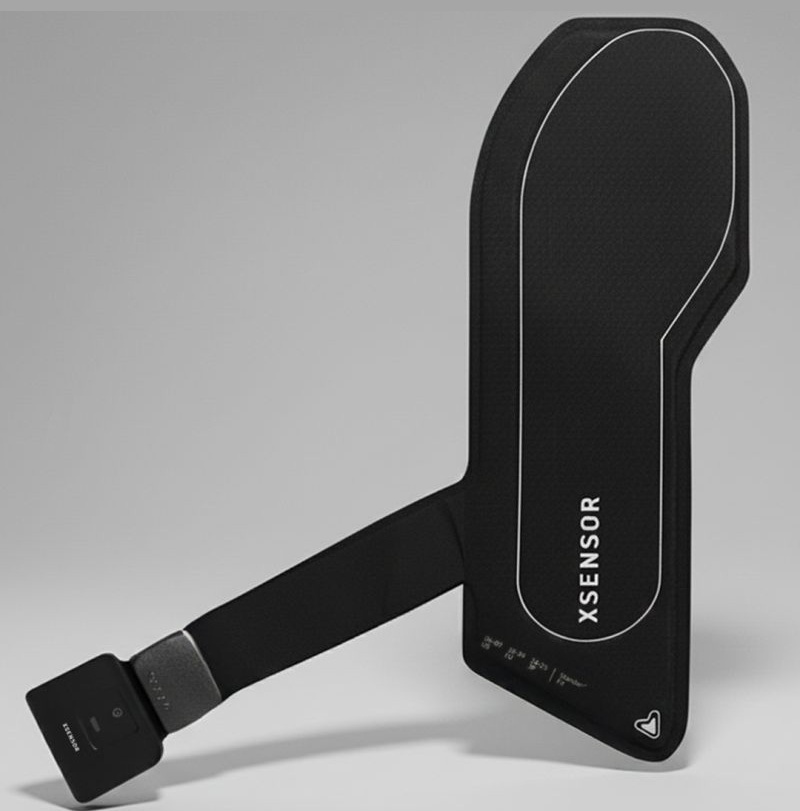}
            \label{fig:xsensor}
    \end{subfigure}%
    \begin{subfigure}[b]{0.4\linewidth}
        \centering
        \includegraphics[width=0.9\textwidth]{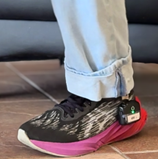}
        \label{fig:xsensor_on}
    \end{subfigure}
    \caption{XSENSOR Insole and IMU System}\label{fig:xsensor_pics}
    \vspace{-4mm}
\end{figure}

\section{Gait Event Detection}
\label{sec:appendix-methods}
\subsection{Transition Duration Computation}
\subsubsection{Sit-to-Stand and Stand-to-Sit (STS)}To extract STS and turning durations from monocular video, we first computed a set of biomechanically meaningful signals from GVHMR joint trajectories. Hip and shoulder midpoints were used to derive vertical and anterior–posterior motion, and a trunk angle signal was computed from the vector connecting the hip midpoint to the shoulder midpoint. These raw trajectories were differentiated to obtain velocities and then smoothed using a Gaussian kernel (\(\mathbf{\sigma}\) = 3) and a 4th‑order low‑pass Butterworth filter ($f_s$ = 30Hz, $f_c$ = 2 Hz) to reduce frame‑to‑frame jitter inherent in video‑based pose estimation. These velocities were utilized to create a weighted composite signal (\text{STS}) defined in \cref{eq:sts}. STS‑1 (sit‑to‑stand) was identified as the largest positive peak, and STS‑2 (stand‑to‑sit) as the largest negative peak. Weights were assigned based on the presumed biomechanical contribution of each feature to sit‑to‑stand performance, with vertical hip displacement (1.0) reflecting primary task completion, anterior shoulder translation (0.7) indexing forward momentum generation, and trunk flexion angle (0.5) capturing compensatory strategy rather than core task success. The findpeaks function in R provided start frame, peak, and end frame for each point. Parameters for peak detection were tuned to each participant and trial, such that minimum peak height was defined as $\overline{STS} \pm 0.8*\mathbf{\sigma_{STS}}$ and minimum peak distance as $0.7 * (Frame_{max}-Frame_{min})$.

\begin{figure*}
    \centering
    \includegraphics[width=0.8\linewidth]{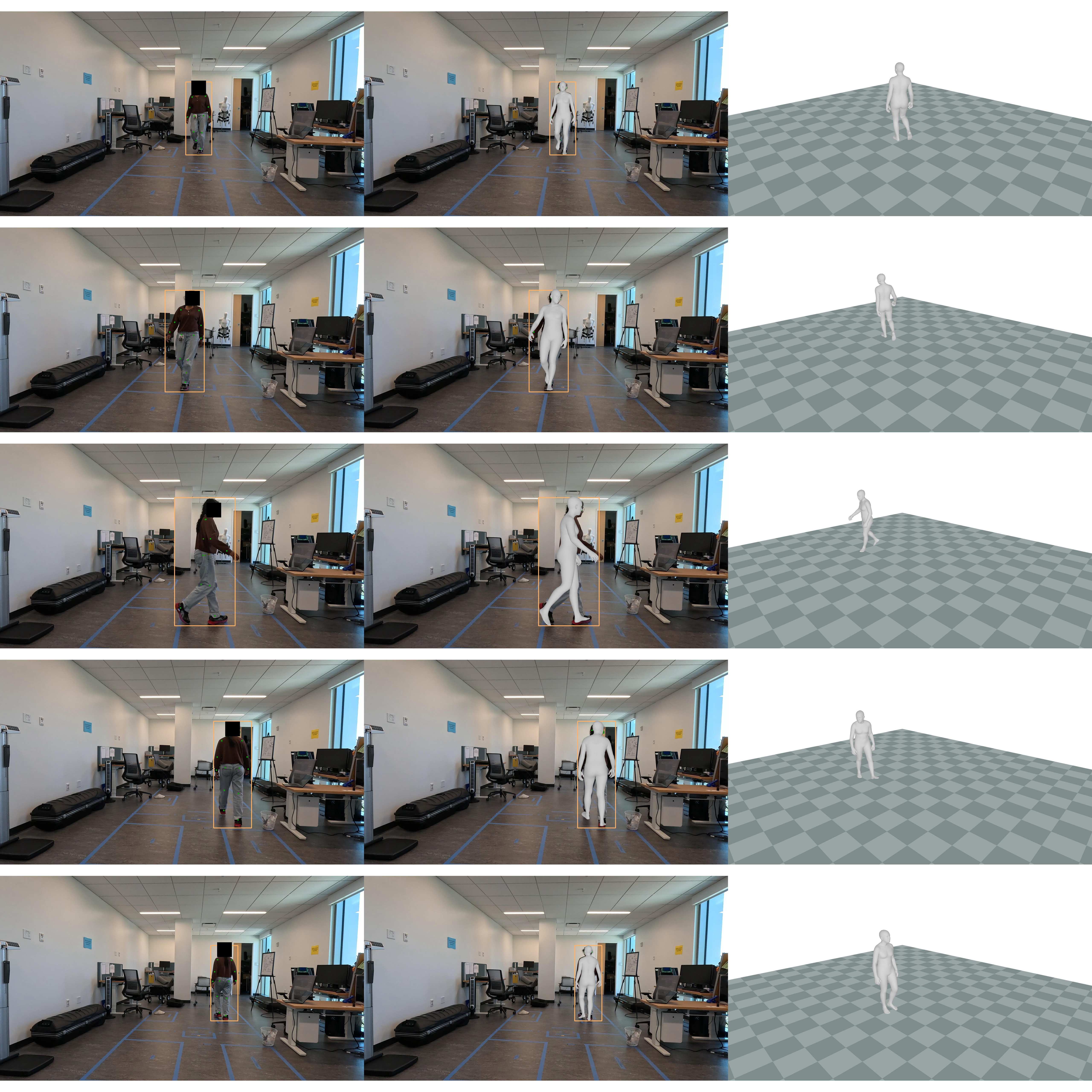}
    \vspace{-2mm}
    \caption{Snapshots of research assistant completing the TUG (left), with the camera-centric view overlay of the HMR (middle), and the global-centric view (right).}
    \label{fig:gait_trajectory1}
    \vspace{-4mm}
\end{figure*}

\begin{figure*}
    \centering
    \includegraphics[width=0.8\linewidth]{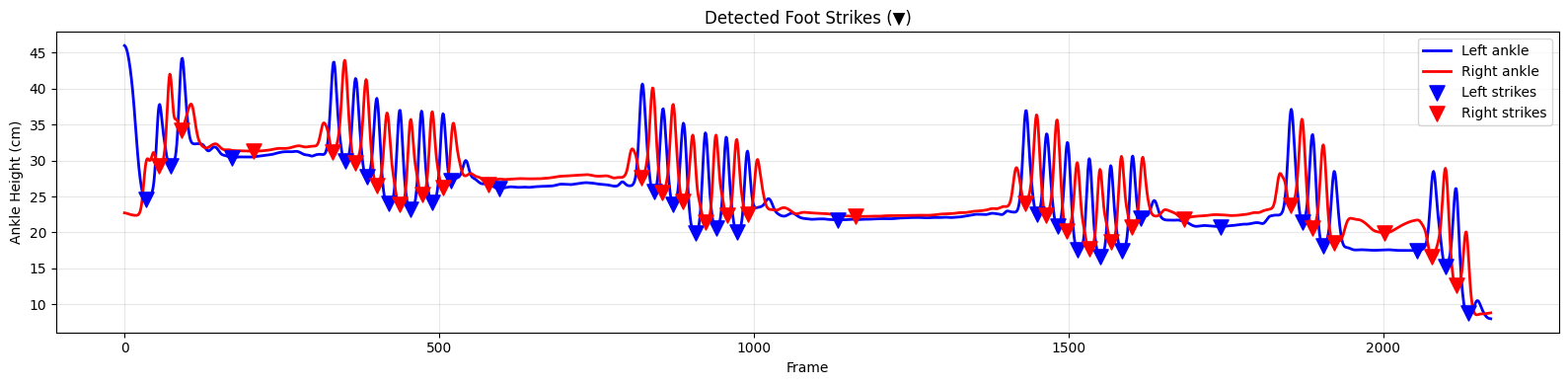}
    \vspace{-2mm}
    \caption{Detection of heel strike using peak detection, according to the detected joints using GVHMR.}
    \label{fig:gait_step_detection}
    \vspace{-4mm}
\end{figure*}

\subsubsection{Turns}
Turning was detected using the filtered hip-line signal, as defined in \cref{eq:hipline}. This hip line signal was differentiated to extract rotational motion of the pelvis. After smoothing with the Gaussian kernel and 4th-order low-pass Butterworth filter, as in the STS signal, $Turn_1$ is defined as a peak and $Turn_2$ is defined as a trough. While the minimum peak height was not changed, the number of peaks was defined as 1 for each trial to limit erroneous peak detection (see \cref{fig:subtask}).

\section{LME Models}
Linear mixed effects (LME) models were utilized in this study to account for the repeated measures model of the data. \(Y_{ij}\) is the outcome for participant \(i\) on trial \(j\), \(\beta_0\) is the fixed intercept, \(\mathbf{x}_{ij}\boldsymbol{\beta}\) represents the fixed effects (fall risk factors and age) and \(b_{0i}\) is the random intercept of participant \(i\), to capture individual-level differences. The error residuals are represented by \(\varepsilon_{ij}\).
LMEs were constructed in R using the \texttt{lmer} function from the \texttt{lme4} library\cite{bates2016}. 
The LME output has several measures in addition to the estimate ($\beta$) and the $p$-value. The full output is presented in the tables below. The estimate is shown alongside the 95\% confidence interval, and  the $p$-value. Below these are the random effects evaluating between- and within-participant variance, model structure, input, and fit.

\begin{equation}
\label{eq:lme}
Y_{ij} = \beta_0 + \mathbf{x}_{ij}\boldsymbol{\beta} + \mathbf{Age}_{ij}\boldsymbol{\beta} + b_{0i} + \varepsilon_{ij}
\end{equation}

$\sigma^2$ refers to the within-participant variability; for instance, the amount of variability of step length trial-to-trial when accounting for the fixed effects (Age and a Fall Risk Factor) in the model. $\tau_{00}$ refers to the between-participant variability; for instance, how variable step lengths are between different participant after accounting for the fixed effects. ICC, or the intra-class correlation coefficient, quantifies the proportion of total variance attributable to differences between participants. $N$ refers to the total number of observations in the clustered model, this may vary depending on missing data for each fall risk factor. In models involving the STEADI, missing data let to a decreased number of participants. Observations represents total number of trials utilized in the model. 
The $R^2$ values provide measure of model fit. The Marginal $R^2$ represents the variance explained only by the fixed effects—in this case, STEADI and age. Marginal $R^2$ explains how much of the outcome (e.g. step length) can be predicted without considering participant‑level differences. Conditional $R^2$ is the variance explained by the full model, including both fixed effects and random effects. Higher values indicates greater model fit, and greater gap between the two $R^2$ values reveals if outcome variability is due to stable, between‑person differences, not trial‑to‑trial fluctuations.


\section{Limitations}

This pilot study is limited by a relatively small sample of 207 videos from 52 older adults from an ongoing study, which may constrain generalizability. Nevertheless, the dense within-participant data, spanning multiple trials and complementary measures including insole-derived step times, self-rated fall risk, fear of falling, and postural sway, strengthen the interpretability of the video-based metrics. Additionally, 18 participants completed assessments at two timepoints at least two months apart, which were treated as independent observations given the temporal separation.

\begin{table*}[t]
\centering
\resizebox{\textwidth}{!}{%
    \begin{tabular}{l rcc rcc rcc}
    \toprule
     & \multicolumn{3}{c}{Model 1} & \multicolumn{3}{c}{Model 2} & \multicolumn{3}{c}{Model 3} \\
    \cmidrule(lr){2-4} \cmidrule(lr){5-7} \cmidrule(lr){8-10}
    Predictors & Estimates & CI & p & Estimates & CI & p & Estimates & CI & p \\
    \midrule
    (Intercept) & 49.74 & [32.01,\ 67.46] & $\mathbf{< 0.001}$ & 65.08 & [48.45,\ 81.72] & $\mathbf{< 0.001}$ & 65.69 & [47.95,\ 83.44] & $\mathbf{< 0.001}$ \\
    STEADI      & -1.36 & [-2.03,\ -0.68] & $\mathbf{< 0.001}$ & ---   & ---             & ---      & ---   & ---             & ---      \\
    Age         & -0.00 & [-0.25,\ \ 0.25] & 0.988   & -0.14 & [-0.37,\ \ 0.10] & 0.264   & -0.23 & [-0.48,\ -0.01] & 0.065    \\
    Short FES-I & ---   & ---             & ---      & -1.04 & [-1.65,\ -0.43] & $\mathbf{0.001}$    & ---   & ---             & ---      \\
    BTrackS     & ---   & ---             & ---      & ---   & ---             & ---      & -0.11 & [-0.27,\ -0.05] & 0.176    \\
    \midrule
    \multicolumn{10}{l}{Random Effects} \\
    \midrule
    $\sigma^2$  & \multicolumn{3}{c}{10.7} & \multicolumn{3}{c}{10.66} & \multicolumn{3}{c}{10.66} \\
    $\tau_{00}$ & \multicolumn{3}{c}{45.1} & \multicolumn{3}{c}{49.72} & \multicolumn{3}{c}{57.08} \\
    ICC         & \multicolumn{3}{c}{0.81} & \multicolumn{3}{c}{0.82}  & \multicolumn{3}{c}{0.84} \\
    N           & \multicolumn{3}{c}{66}   & \multicolumn{3}{c}{70}    & \multicolumn{3}{c}{70} \\
    \midrule
    Observations      & \multicolumn{3}{c}{195} & \multicolumn{3}{c}{207} & \multicolumn{3}{c}{207} \\
    Marginal R$^2$    & \multicolumn{3}{c}{0.210} & \multicolumn{3}{c}{0.180} & \multicolumn{3}{c}{0.084} \\
    Conditional R$^2$ & \multicolumn{3}{c}{0.848} & \multicolumn{3}{c}{0.855} & \multicolumn{3}{c}{0.856} \\
    \bottomrule
    \end{tabular}%
}
\caption{Step Length (SL) Analysis}
\label{tab:sl_full}
\end{table*}

\begin{table*}[t]
\centering
\resizebox{\textwidth}{!}{%
    \begin{tabular}{l rcc rcc rcc}
    \toprule
     & \multicolumn{3}{c}{Model 1} & \multicolumn{3}{c}{Model 2} & \multicolumn{3}{c}{Model 3} \\
    \cmidrule(lr){2-4} \cmidrule(lr){5-7} \cmidrule(lr){8-10}
    Predictors & Estimates & CI & p & Estimates & CI & p & Estimates & CI & p \\
    \midrule
    (Intercept)              & 296.47 & [160.02,\ 432.92] & $\mathbf{<0.001}$ & 301.74 & [182.35,\ 421.13] & $\mathbf{<0.001}$ & 302.56 & [182.97,\ 422.14] & $\mathbf{<0.001}$ \\
    STEADI                   & -0.14  & [-5.34,\ \ 5.05]  & 0.956             & ---    & ---               & ---               & ---    & ---               & ---               \\
    Age                      & -2.79  & [-4.73,\ -0.86]   & $\mathbf{0.005}$  & -2.72  & [-4.43,\ -1.02]   & $\mathbf{0.002}$  & -2.95  & [-4.61,\ -1.29]   & $\mathbf{0.001}$  \\
    Short FES-I              & ---    & ---               & ---               & -1.24  & [-5.61,\ \ 3.13]  & 0.577             & ---    & ---               & ---               \\
    BTrackS    & ---    & ---               & ---               & ---    & ---               & ---               & 0.14   & [-0.93,\ \ 1.22]  & 0.794             \\
    \midrule
    \multicolumn{10}{l}{Random Effects} \\
    \midrule
    $\sigma^2$        & \multicolumn{3}{c}{960.87}  & \multicolumn{3}{c}{949.11}  & \multicolumn{3}{c}{949.11}  \\
    $\tau_{00}$       & \multicolumn{3}{c}{2563.93} & \multicolumn{3}{c}{2426.50} & \multicolumn{3}{c}{2436.54} \\
    ICC               & \multicolumn{3}{c}{0.73}    & \multicolumn{3}{c}{0.72}    & \multicolumn{3}{c}{0.72}    \\
    N                 & \multicolumn{3}{c}{66}      & \multicolumn{3}{c}{70}      & \multicolumn{3}{c}{70}      \\
    \midrule
    Observations      & \multicolumn{3}{c}{195}   & \multicolumn{3}{c}{207}   & \multicolumn{3}{c}{207}   \\
    Marginal R$^2$    & \multicolumn{3}{c}{0.117} & \multicolumn{3}{c}{0.133} & \multicolumn{3}{c}{0.130} \\
    Conditional R$^2$ & \multicolumn{3}{c}{0.759} & \multicolumn{3}{c}{0.756} & \multicolumn{3}{c}{0.756} \\
    \bottomrule
    \end{tabular}%
}
\caption{SW Variability Analysis}
\label{tab:sw_variability}
\end{table*}

\begin{table*}[t]
\centering
\resizebox{\textwidth}{!}{%
    \begin{tabular}{l rcc rcc rcc}
    \toprule
     & \multicolumn{3}{c}{Model 1} & \multicolumn{3}{c}{Model 2} & \multicolumn{3}{c}{Model 3} \\
    \cmidrule(lr){2-4} \cmidrule(lr){5-7} \cmidrule(lr){8-10}
    Predictors & Estimates & CI & p & Estimates & CI & p & Estimates & CI & p \\
    \midrule
    (Intercept)           & 393.99 & [110.09,\ 677.89]  & $\mathbf{0.007}$  & 594.80  & [330.99,\ 858.61]  & $\mathbf{<0.001}$ & 602.10  & [330.27,\ 873.93]  & $\mathbf{<0.001}$ \\
    STEADI                & -19.62 & [-30.44,\ -8.80]   & $\mathbf{<0.001}$ & ---     & ---                & ---               & ---     & ---                & ---               \\
    Age                   & -0.59  & [-4.61,\ \ 3.43]   & 0.772             & -2.50   & [-6.27,\ \ 1.27]   & 0.193             & -3.44   & [-7.21,\ \ 0.34]   & 0.074             \\
    Short FES-I           & ---    & ---                & ---               & -12.76  & [-22.43,\ -3.09]   & $\mathbf{0.010}$  & ---     & ---                & ---               \\
    BTrackS Balance Score & ---    & ---                & ---               & ---     & ---                & ---               & -1.97   & [-4.42,\ \ 0.47]   & 0.113             \\
    \midrule
    \multicolumn{10}{l}{Random Effects} \\
    \midrule
    $\sigma^2$        & \multicolumn{3}{c}{11382.63} & \multicolumn{3}{c}{11061.41} & \multicolumn{3}{c}{11055.43} \\
    $\tau_{00}$       & \multicolumn{3}{c}{8672.18}  & \multicolumn{3}{c}{9695.94}  & \multicolumn{3}{c}{10532.24} \\
    ICC               & \multicolumn{3}{c}{0.43}     & \multicolumn{3}{c}{0.47}     & \multicolumn{3}{c}{0.49}     \\
    N                 & \multicolumn{3}{c}{66}       & \multicolumn{3}{c}{70}       & \multicolumn{3}{c}{70}       \\
    \midrule
    Observations      & \multicolumn{3}{c}{195}   & \multicolumn{3}{c}{207}   & \multicolumn{3}{c}{207}   \\
    Marginal R$^2$    & \multicolumn{3}{c}{0.143} & \multicolumn{3}{c}{0.103} & \multicolumn{3}{c}{0.069} \\
    Conditional R$^2$ & \multicolumn{3}{c}{0.513} & \multicolumn{3}{c}{0.522} & \multicolumn{3}{c}{0.523} \\
    \bottomrule
    \end{tabular}%
}
\caption{SL Variability Analysis}
\label{tab:sl_variability}
\end{table*}

\begin{table*}[t]
\centering
\resizebox{\textwidth}{!}{%
    \begin{tabular}{l rcc rcc rcc}
    \toprule
     & \multicolumn{3}{c}{Model 1} & \multicolumn{3}{c}{Model 2} & \multicolumn{3}{c}{Model 3} \\
    \cmidrule(lr){2-4} \cmidrule(lr){5-7} \cmidrule(lr){8-10}
    Predictors & Estimates & CI & p & Estimates & CI & p & Estimates & CI & p \\
    \midrule
    (Intercept)           & 26.82 & [5.86,\ \ 47.79]  & $\mathbf{0.012}$ & 13.39 & [-7.16,\ \ 33.95] & 0.200            & 13.69 & [-7.46,\ \ 34.83] & 0.203            \\
    STEADI                & 1.23  & [0.45,\ \ \ 2.01] & $\mathbf{0.002}$ & ---   & ---               & ---              & ---   & ---               & ---              \\
    Age                   & 0.21  & [-0.08,\ \ 0.51]  & 0.157            & 0.36  & [0.07,\ \ \ 0.65] & $\mathbf{0.016}$ & 0.44  & [0.15,\ \ \ 0.74] & $\mathbf{0.003}$ \\
    Short FES-I           & ---   & ---               & ---              & 0.71  & [-0.01,\ \ 1.42]  & 0.052            & ---   & ---               & ---              \\
    BTrackS Balance Score & ---   & ---               & ---              & ---   & ---               & ---              & 0.00  & [-0.19,\ \ 0.19]  & 0.989            \\
    \midrule
    \multicolumn{10}{l}{Random Effects} \\
    \midrule
    $\sigma^2$  & \multicolumn{3}{c}{172.17}                        & \multicolumn{3}{c}{175.44}                        & \multicolumn{3}{c}{175.51}                        \\
    $\tau_{00}$ & \multicolumn{3}{c}{15.18} & \multicolumn{3}{c}{21.18} & \multicolumn{3}{c}{25.47} \\
    ICC         & \multicolumn{3}{c}{0.08}                          & \multicolumn{3}{c}{0.11}                          & \multicolumn{3}{c}{0.13}                          \\
    N           & \multicolumn{3}{c}{71}    & \multicolumn{3}{c}{73}    & \multicolumn{3}{c}{73}    \\
    \midrule
    Observations      & \multicolumn{3}{c}{211}   & \multicolumn{3}{c}{217}   & \multicolumn{3}{c}{217}   \\
    Marginal R$^2$    & \multicolumn{3}{c}{0.092} & \multicolumn{3}{c}{0.071} & \multicolumn{3}{c}{0.051} \\
    Conditional R$^2$ & \multicolumn{3}{c}{0.165} & \multicolumn{3}{c}{0.171} & \multicolumn{3}{c}{0.172} \\
    \bottomrule
    \end{tabular}%
}
\caption{Sit-to-Stand and Stand-to-Sit (STS) Duration Analysis}
\label{tab:sts_durations}
\end{table*}

\end{document}